\documentclass[conference]{IEEEtran}
\IEEEoverridecommandlockouts

\usepackage{generic}
\usepackage{cite}
\usepackage{amsmath,amssymb,amsfonts}

\usepackage{graphicx}

\usepackage{hyperref}
\usepackage{textcomp}

\usepackage{graphicx}
\usepackage{dcolumn}
\usepackage{bm}
\usepackage{multirow}
\usepackage{xcolor}
\usepackage{braket}
\usepackage{algorithm}
\usepackage{algpseudocode}
\usepackage{hyperref,url}
\usepackage{mathtools}
\usepackage{csquotes}
\usepackage{lineno}
\usepackage{float}
\usepackage{tabularx}
\usepackage{booktabs}
\usepackage{threeparttable}
\usepackage{lipsum} 
\usepackage{comment}

\begin{document}


\title{ Multimodal Quantum Vision Transformer for Enzyme Commission Classification from Biochemical Representations}

\author{%

    \IEEEauthorblockN{Murat Isik}
    \IEEEauthorblockA{\textit{Purdue University} \\
    West Lafayette, USA \\
    misik@purdue.edu \\
    }
    \and
    \IEEEauthorblockN{Mandeep Kaur Saggi}
    \IEEEauthorblockA{\textit{NC State University} \\
    Raleigh, USA \\
    mksaggi@ncsu.edu \\
    }
    \and
    \IEEEauthorblockN{Humaira Gowher}
    \IEEEauthorblockA{\textit{Purdue University} \\
    West Lafayette, USA \\
    hgowher@purdue.edu}
    
\and
    
    \IEEEauthorblockN{Sabre Kais}
    \IEEEauthorblockA{\textit{NC State University} \\
    Raleigh, USA \\
    skais@ncsu.edu}
        \and

 } 

 \maketitle

\begin{abstract} Accurately predicting enzyme functionality remains one of the major challenges in computational biology, particularly for enzymes with limited structural annotations or sequence homology. We present a novel multimodal Quantum Machine Learning (QML) framework that enhances Enzyme Commission (EC) classification by integrating four complementary biochemical modalities: protein sequence embeddings, quantum-derived electronic descriptors, molecular graph structures, and 2D molecular image representations. Quantum Vision Transformer (QVT) backbone equipped with modality-specific encoders and a unified cross-attention fusion module. By integrating graph features and spatial patterns, our method captures key stereoelectronic interactions behind enzyme function. Experimental results demonstrate that our multimodal QVT model achieves a top-1 accuracy of 85.1\%, outperforming sequence-only baselines by a substantial margin and achieving better performance results compared to other QML models. \end{abstract}


\section{Introduction}
Enzymes play a pivotal role in virtually every aspect of biological chemistry, acting as highly specialized catalysts that drive metabolic pathways, DNA replication, cell signaling, and other essential processes in living organisms \cite{tripathi2020industrial, shoda2016enzymes}. Their remarkable efficiency and specificity stem from the intricate interplay of stereoelectronic features, thermodynamic properties, and structural conformations within active sites. Consequently, the ability to accurately predict enzyme function has immense implications, facilitating the discovery of novel biocatalysts, guiding metabolic engineering efforts, and accelerating drug development. However, this seemingly straightforward task of predicting enzymatic roles from primary sequence information is fraught with complexity because of the multifaceted nature of enzyme catalysis and the subtle biochemical cues that underlie functional specificity \cite{luo2025learning, thurimella2024harnessing}.

Early approaches to enzyme classification often relied on sequence alignment techniques and manually curated similarity metrics. These methods proved effective for well-studied enzymes with abundant homologs or highly conserved active site motifs, but their performance diminished for more divergent sequences or newly discovered enzymes with scarce structural annotations. As a result, computational biologists began exploring statistical and machine learning (ML) methods that could capture higher-order sequence relationships, motif conservation, and evolutionary signals \cite{sampaio2023machine}, \cite{salas2024machine}. Although these techniques offered an improvement over purely alignment-based strategies, many of them still struggled to represent three-dimensional folds, molecular chirality, and local electronic environment factors that critically determine how enzymes bind substrates and stabilize transition states. Over the past decade, deep learning architectures have begun to address certain limitations by extracting context-rich embeddings from protein sequences \cite{catacutan2024machine}, \cite{kyro2025t}. Convolutional and recurrent neural networks, followed by transformer-based models, have shown promise in decoding long-range interactions and subtle biochemical semantics. In tandem with deep learning progress, the field of quantum machine learning (QML) has garnered increasing attention. By tapping into principles of quantum mechanics and exploiting quantum-enhanced descriptors, such as self-consistent field (SCF) energies, nuclear repulsion terms, and molecular orbital distributions, QML approaches hold the potential to capture phenomena that remain elusive to purely classical methods. When paired with modern neural architectures, these quantum-derived features shed light on the electron density shifts and non-covalent interactions that define catalytic activity \cite{luo2022understanding}.

To leverage the complementary strengths of modern deep learning and quantum computational insights, we propose a multimodal Quantum Vision Transformer (QVT) framework. Our goal is to address the limitations of sequence-only models by incorporating four distinct yet interrelated modalities: (1) protein sequence embeddings, (2) quantum-derived electronic descriptors, (3) graph-based molecular representations, and (4) 2D molecular image data. This fusion of biochemical perspectives enables the model to capture diverse layers of information, from residue-level evolutionary patterns to global molecular topologies and electronic configurations.

\section{Background}
Research on enzyme function prediction has long been at the crossroads of computational biology and machine learning, with recent advances highlighting the importance of integrating diverse data sources. In classical approaches, computational biologists primarily relied on sequence similarity and motif searches, often supported by well-curated databases of evolutionary relationships. While these techniques provided critical insights for families of enzymes sharing highly conserved domains, their performance declined substantially when attempting to classify novel or poorly annotated protein sequences \cite{ryu2019deep}, \cite{li2018deepre}.

\subsection{Quantum Machine Learning (QML)}

By leveraging computational principles rooted in quantum mechanics, such as superposition and entanglement, QML methods can encode quantum-chemical descriptors with higher fidelity than their purely classical counterparts \cite{biamonte2017quantum}. Early work in QML demonstrated that machine learning models could, for instance, predict molecular atomization energies and electronic properties with notable accuracy by ingesting quantum-computed features \cite{rupp2012fast, montavon2013machine}. Although much of this pioneering effort focused on small molecules and materials, recent advances have laid the groundwork for applying QML to protein systems, including enzymatic complexes with bound substrates or cofactors. QML methods attempt to tackle these challenges by introducing quantum mechanical principles into the modeling process. Techniques such as Quantum Variational Circuits (QVC), QVT, and Quantum Support Vector Machines (QSVM) exploit quantum computing’s intrinsic capacity to encode and manipulate high-dimensional data, potentially leading to enhanced recognition of subtle electron density correlations or transition state formations \cite{damborsky2024accelerating}. 

\subsection{Enzyme Commission (EC) Prediction from Sequences}

The most direct route to enzyme function prediction remains sequence-based annotation, where each amino acid sequence is inspected for key motifs, functional residues, and structural domains. Deep learning architectures, particularly convolutional and transformer-based models, have shown promise by mining long-range interactions and context-specific patterns \cite{mienye2024comprehensive}. Recent research highlights the efficacy of encoding protein sequences into high-dimensional representations before feeding them into classification networks, thereby capturing subtle biochemical signatures such as post-translational modification sites or binding pockets \cite{villalobos2022protein}. Nonetheless, relying exclusively on linear sequences can overlook critical structural and electronic factors that can be especially acute in complex enzymes with flexible conformations or intricate active sites.

\subsection{Image- and Graph-based Approaches}

To better account for spatial and topological features, image- and graph-based representations have gained attention within the computational chemistry and drug design communities. Graph neural networks (GNNs), for example, transform molecular structures into nodes and edges, enabling the model to learn from connectivity patterns and local substructures \cite{gilmer2017neural, coley2019graph}. This approach has proven especially useful for capturing the influence of aromatic rings, charged functional groups, and stereochemistry on catalytic activities and binding affinities. Meanwhile, image-based frameworks convert molecular structures into 2D renderings, which can then be processed by convolutional neural networks (CNNs) to identify salient visual cues such as ring systems, branching patterns, and potential sites of reactivity \cite{chandrasekaran2021image}. While these methods lose some three-dimensional detail, they often excel at distinguishing closely related chemical structures. For enzyme tasks, 2D images of substrates or cofactors complement protein sequence data, providing a more complete view of biochemical function. Increasingly, researchers recognize that each data modality, whether sequence, quantum-derived descriptor, graph, or image, captures distinct aspects of enzyme functionality \cite{baltruvsaitis2018multimodal}. Protein sequences capture evolutionary and functional information at the residue level, while quantum descriptors reveal critical electronic properties driving catalysis.

\begin{table}[h]
    \centering
    \footnotesize 
    \caption{Overview of different data types in the dataset.}
    \resizebox{0.45\textwidth}{!}{
    \begin{tabularx}{0.5\textwidth}{l>{\raggedright\arraybackslash}X>{\raggedright\arraybackslash}X}
        \toprule
        \textbf{Data Type} & \textbf{Description} & \textbf{Applications} \\
        \midrule
        Protein Sequences & Amino acid sequences mapped to UniProt IDs, cleaned and aligned. & Capturing evolutionary patterns, domains, and motifs. \\
        Graph Representations & Molecular graphs from SMILES strings, atom connectivity and bond info. & Highlighting substructures influencing enzymatic activity. \\
        Image Representations & 2D molecular images of substrates and products via RDKit. & Identifying rings, branches, and stereochemistry. \\
        Quantum Descriptors & SCF energy, nuclear repulsion, and gradient magnitudes via QM/MM. & Capturing stereoelectronic effects on catalysis. \\
        EC Numbers (Labels) & Enzyme Commission numbers linked to catalytic functions. & Providing labels for supervised classification. \\
        \bottomrule
    \end{tabularx}}
    \label{tab:dataset_overview}
\end{table}

\vspace{-7pt}

\section{Methodology}

\subsection{Dataset} 
Our dataset was compiled by aggregating multiple biochemical sources to capture a diverse range of enzyme classes and relevant molecular information. Specifically, we collected data from the following repositories:

\begin{itemize}
\item \textbf{RHEA} \cite{morgat2016updates}: An expert-curated database of biochemical reactions, used to obtain high-quality reaction and enzyme annotations. 
\item \textbf{UniProt} \cite{pundir2017uniprot}: A comprehensive resource of protein sequence and functional information, from which we extracted amino acid sequences and corresponding UniProt IDs. 
\end{itemize}

From these sources, we curated five key data modalities for our enzyme function prediction task:

\begin{itemize}
\item \textbf{Protein Sequences}: Cleaned and aligned amino acid sequences mapped to UniProt IDs. These sequences serve as the primary input for capturing evolutionary and functional patterns. 
\item \textbf{EC Numbers}: Target labels describing catalytic activity, enabling supervised training for enzyme classification \cite{probst2022biocatalysed}. 
\item \textbf{SMILES and SELFIES Representations}: Canonical Simplified Molecular Input Line Entry System (SMILES) strings and their Self-Referencing Embedded Strings (SELFIES) equivalents were tokenized and one-hot encoded, enabling natural language processing techniques for chemical text data \cite{lo2023recent}. This dual representation captures chemical robustness and error correction in sequence models.

\item \textbf{Graph Representations}: Molecular graphs generated from SMILES strings using Open-source RDKit toolkit \cite{landrum2019rdkit}, encompassing atom connectivity and bond information. 
\item \textbf{Image Representations}: 2D structural visualizations of enzyme substrates and products, also generated via RDKit. These images aid in identifying visually salient features such as ring systems and branching. 
\item \textbf{Quantum Descriptors}: SCF Total Energy, Nuclear Repulsion Energy, and Gradient Magnitudes computed via Q-Chem \cite{Qchem, shao2015advances, gugler2022quantum} in QM/MM setups. These descriptors supply a rich electronic signature that classical features alone may fail to capture.
\end{itemize}

The dataset encompasses five distinct modalities, each contributing complementary biochemical insights for the enzyme classification task, as summarized in \autoref{tab:dataset_overview}. Data preprocessing included standardization of amino acid sequences, removal of ambiguous or low-quality entries, and normalization of quantum descriptors. For image and graph representations, we verified chemical correctness of SMILES strings to avoid spurious molecular structures.

\begin{figure}[ht]
\centering
\includegraphics[width=0.48\textwidth]{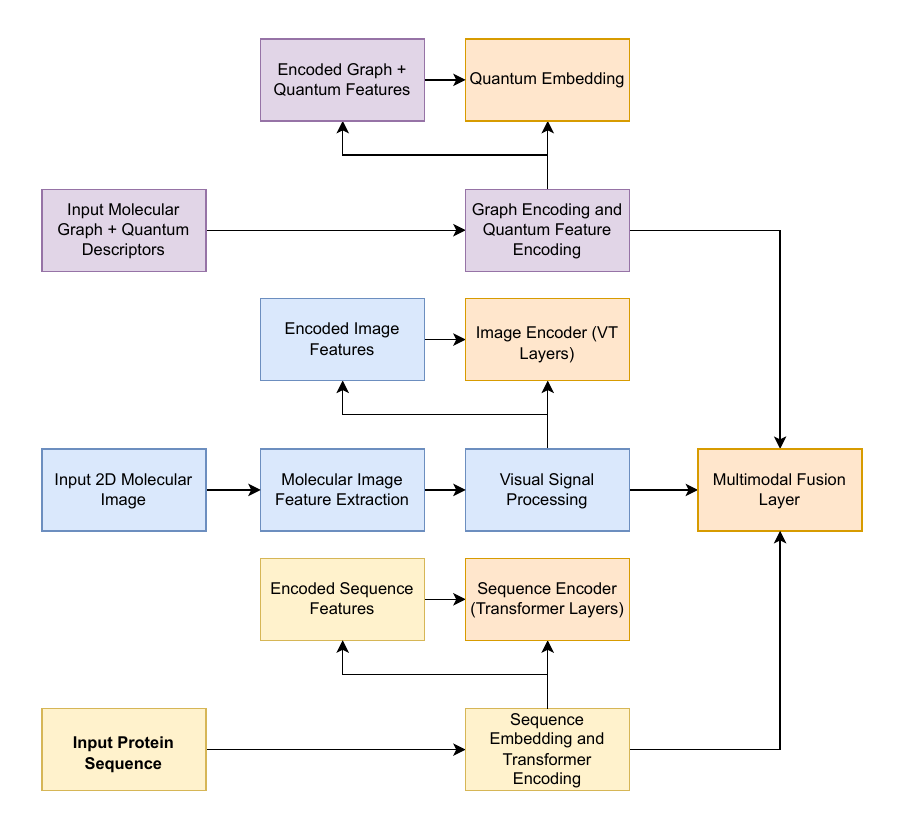}
\caption{Data processing diagram for the Quantum Vision Transformer (QVT) framework. Protein sequences, 2D molecular images, and molecular graphs augmented with quantum descriptors are independently processed through modality-specific encoders and then fused for enzyme function prediction.}
\label{fig:data}
\end{figure}

\autoref{fig:data} outlines the data processing workflow for the QVT framework. The pipeline begins with three distinct input modalities: protein sequences, 2D molecular images, and molecular graphs augmented with quantum descriptors. Protein sequences undergo embedding and transformer encoding to extract sequence features. Molecular images are processed through a Vision Transformer (VT)-style encoder to derive visual feature representations. Meanwhile, molecular graphs and quantum descriptors are jointly encoded into a unified feature space. These modality-specific features, sequence embeddings, image embeddings, and graph-quantum embeddings are then integrated via a multimodal fusion layer, enabling comprehensive biochemical representation learning for downstream enzyme function prediction.

\subsection{Quantum Vision Transformer (QVT) Architecture} We develop a multimodal QVT that draws upon quantum-derived descriptors, protein sequence embeddings, 2D molecular images, and graph-based structures. This design is aimed at exploiting complementary biochemical information and better reflecting the underlying catalytic mechanisms.

\subsubsection{Quantum-Derived Features} Quantum mechanical calculations offer insights into electronic properties that may directly influence catalytic activity. In our framework, we incorporate SCF Total Energy, Nuclear Repulsion Energy, and optimized gradient magnitudes, all derived from Q-Chem simulations \cite{Qchem}. These values capture key stereoelectronic and energetic characteristics relevant to catalysis. For integration into the QVT model, the computed descriptors were normalized and amplitude-encoded into quantum states using a Möttönen state \cite{mottonen2004quantum} preparation routine. All quantum state preparations and circuit executions were performed on IBM’s Qiskit AerSimulator backend, ensuring reproducibility and allowing simulation of high-dimensional quantum states beyond current hardware qubit limits. This simulated quantum layer enabled efficient experimentation while preserving the physical meaning of the computed features.

\subsubsection{Protein Sequence Embeddings} Enzyme sequences are tokenized at the amino acid level and processed by a transformer-based encoder. We use an embedding dimension of 1024, enabling the capture of long-range dependencies critical to enzyme functionality. Multi-head self-attention layers preserve contextual clues across distant residues, thus modeling how variations in sequence composition influence catalytic efficacy.

\subsubsection{Molecular Graphs} Graph-based representations supply a topological perspective on molecular structures, highlighting chemical bonds and substructures that may affect binding and catalysis. We convert SMILES strings into molecular graphs via RDKit, and then feed them into a GNN module composed of three message-passing layers. Residual connections between layers help maintain stability during training and enable the extraction of features like node centrality, bond order, and subgraph motifs. These properties are often decisive in shaping the enzyme-substrate interface.

\subsubsection{Molecular Images} We also generate 2D molecular depictions for each substrate or product and feed these into a lightweight convolutional encoder to capture visually salient features. Employing a three-layer CNN backbone with ReLU activations, we extract hierarchical image features indicative of ring systems, stereocenters, and other functional groups. While 2D images lose some three-dimensional information, they often highlight critical distinctions in chemical scaffolds that may otherwise be obscured in purely symbolic representations.

\subsubsection{Multimodal Fusion and Classification} After feature extraction, the representations from each modality (quantum, sequence, graph, and image) are concatenated and integrated through a cross-attention fusion layer. This mechanism learns how different channels of biochemical information interact, for example, correlating a particular quantum descriptor with a specific subgraph motif. The fused vector is then passed through fully connected layers with dropout for regularization, and finally into a softmax classifier for EC classification.

\begin{figure}[ht]
\includegraphics[width=0.5\textwidth]{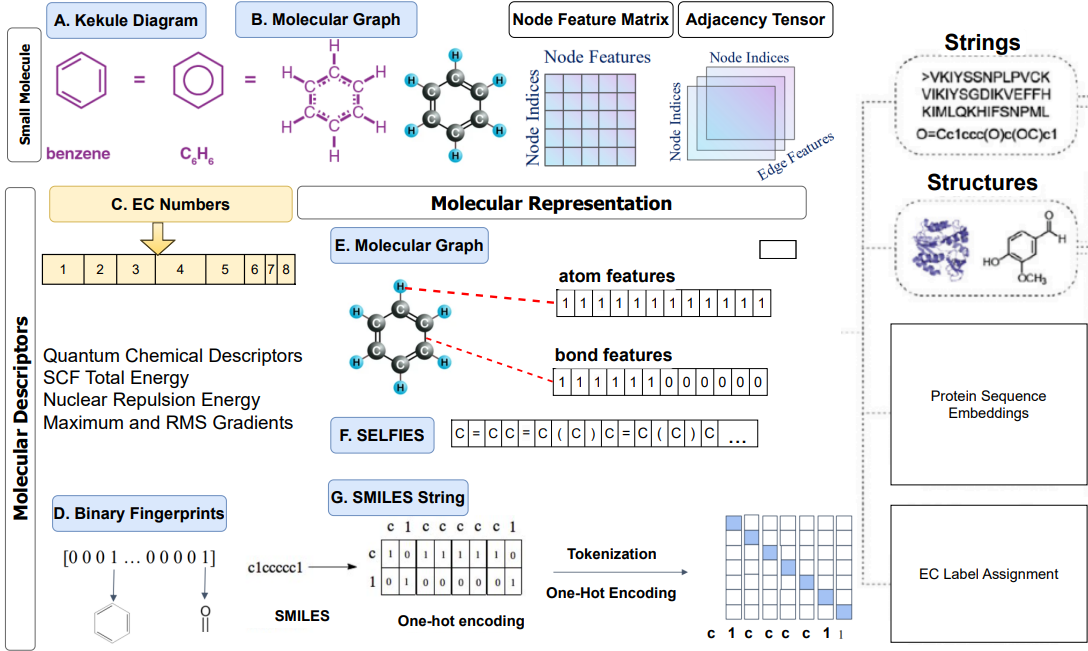}
\caption{Overview of the multimodal Quantum Vision Transformer (QVT) framework for enzyme classification. The pipeline integrates multiple biochemical representations: (A) Kekulé structural diagrams and (B) molecular graph adjacency tensors capture atomic connectivity; (C) Enzyme Commission (EC) numbers provide functional labels for supervised learning; (D) binary fingerprints encode molecular substructures; (E) molecular graphs preserve spatial bonding information; (F) SELFIES representations offer robust string-based encodings; (G) SMILES strings are processed through a sequence embedding pipeline. These data streams are further enriched through natural language processing, chemical descriptor extraction, and quantum chemical descriptor computation. }

\label{fig:qvt}
\end{figure}

\autoref{fig:qvt} presents a comprehensive overview of the data processing and model architecture for the proposed QVT framework. Starting from molecular representations (including Kekulé diagrams, graph adjacency matrices, SMILES strings, and SELFIES sequences), multiple biochemical features are extracted. Binary fingerprints and graph-based embeddings encode atom-level and bond-level information. Additionally, quantum chemical descriptors (self-consistent field total energy, nuclear repulsion energy, and optimized gradient magnitudes) are computed to capture critical electronic properties of enzyme substrates. 2D molecular images are generated to provide spatial and stereoelectronic cues. These diverse data modalities are individually processed through dedicated embedding pipelines, such as natural language processing for sequences and convolutional encoders for images. The multimodal features are subsequently fused via a cross-attention fusion layer to produce integrated molecular embeddings, which are used for multi-class EC classification.

\subsection{Implementation Details} We implemented the QVT architecture using PyTorch, while the quantum routines (amplitude encoding) are prototyped with Qiskit \cite{QiskitGitHub} in a simulator environment for scalability. Model optimization employed the Adam optimizer, paired with a focal loss to address class imbalance, often an issue in enzymology datasets where certain classes (oxidoreductases) dominate the training set. Hyperparameters such as the number of GNN layers, CNN filters, and attention heads in the transformer were fine-tuned based on a combination of grid search and domain knowledge.

To ensure robust evaluation, we partitioned the dataset into training, validation, and test splits using a stratified approach that maintained class ratios. Our primary metric was top-1 accuracy, reflecting the model’s ability to correctly identify the most likely EC number. Ablation studies were conducted by selectively removing one or more data modalities (quantum descriptors and image features) to assess their individual contributions to overall performance. This multi-faceted evaluation strategy underscores how different biochemical perspectives, when combined in a unified pipeline, can yield improved accuracy and richer biological insight.

\begin{figure}
    \hspace*{-0.5cm}  
    \graphicspath{{D:/Stack/}} 
    \vspace{-10pt}\includegraphics[width=0.55\textwidth]{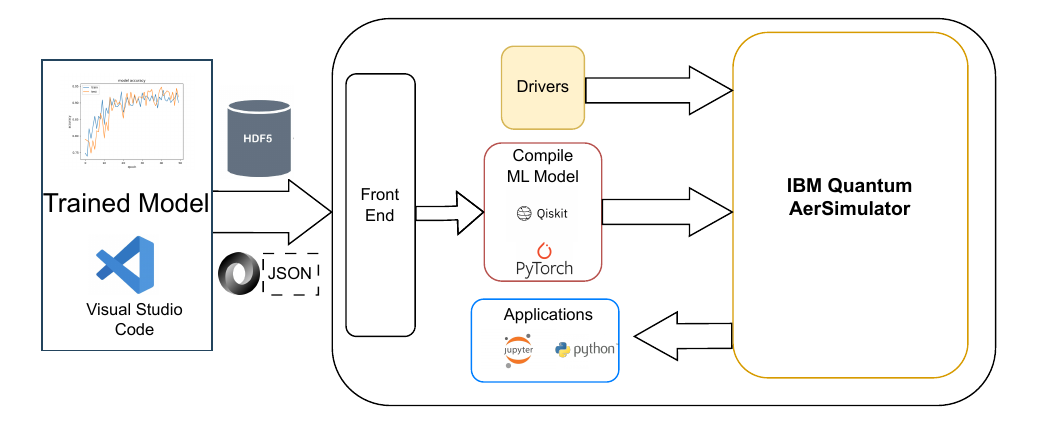}
    \caption{Deployment pipeline of a trained QVT model, highlighting compilation, driver integration, frontend interfacing, and execution on IBM Quantum AerSimulator.}
    \label{test}
\end{figure}

\autoref{test} The diagram outlines the workflow for deploying a trained QML framework. The workflow initiates with the compilation of the trained model, which includes both classical neural components and quantum-derived modules, ensuring compatibility with quantum simulators and hardware backends. Following compilation, hardware-specific drivers are employed, such as Qiskit transpilers and run-time interfaces, to manage circuit execution and resource allocation. The deployment interface is developed using Visual Studio Code, where input/output schemas and interaction protocols are implemented. To enable platform-agnostic communication, data representations (including biochemical inputs and enzyme classification outputs) are serialized using standardized JSON formats. This allows seamless integration with external applications and APIs. The serialized model is then executed on IBM Quantum infrastructure via the AerSimulator.

\begin{table}[h]
\hspace*{-0.5cm} 
\begin{threeparttable}
\caption{Incremental Performance Improvements for Multimodal Quantum Vision Transformer (QVT)}
\begin{tabular}{|l|c|c|c|c|}
\hline
\textbf{Model Variant} & \textbf{Top-1 Acc} & \textbf{Precision} & \textbf{Recall} & \textbf{F1-Score} \\
\hline
SMILES/SELFIES & \textcolor{red}{74.0} & \textcolor{red}{73.2} & \textcolor{red}{72.5} & \textcolor{red}{72.8} \\
+ Quantum Descriptors & & & & \\
\hline
+ Molecular Graphs & 82.4 & 81.1 & 80.5 & 80.8 \\
+ 2D Molecular Images & 84.0 & 83.2 & 82.6 & 82.9 \\
+ Molecular Fingerprints & \textbf{85.1} & \textbf{84.5} & \textbf{83.8} & \textbf{84.1} \\
\hline
\end{tabular}
\label{tab:ablation_incremental}
\begin{tablenotes}
\footnotesize
\item[*] SMILES/SELFIES Quantum Descriptors is baseline model.
\item[*] Metrics are macro-averaged across enzyme classes unless otherwise specified.
\item[**] Performance was evaluated on a stratified test set (20\% split) using the same model hyperparameters and training epochs to ensure fair comparisons. 
\end{tablenotes}
\end{threeparttable}
\end{table}

\section{Results}

Our multimodal QVT framework demonstrated a marked improvement over traditional sequence-only baselines in predicting EC numbers. These results suggest that incorporating complementary modalities such as molecular graphs, 2D images, and quantum descriptors substantially enhances the model’s capacity to differentiate among functionally diverse enzymes.

\autoref{tab:ablation_incremental} summarizes the progressive performance improvements achieved by incrementally incorporating additional biochemical modalities into the QVT framework. The baseline configuration, which leverages SMILES/SELFIES representations combined with quantum descriptors (SCF Total Energy, Nuclear Repulsion Energy), yields a Top-1 accuracy of 74.0\%, along with precision, recall, and F1-score values of 73.2\%, 72.5\%, and 72.8\%, respectively. Augmenting the model with molecular graph representations significantly enhances performance across all metrics boosting accuracy to 82.4\%, and increasing precision, recall, and F1-score to 81.1\%, 80.5\%, and 80.8\%. The addition of 2D molecular images contributes further improvements, particularly in capturing stereochemical and spatial features, resulting in an accuracy of 84.0\% and corresponding increases in precision (83.2\%), recall (82.6\%), and F1-score (82.9\%). Finally, incorporating molecular fingerprints vectorized encodings of substructural patterns leads to the best overall performance, with the model achieving 85.1\% accuracy, 84.5\% precision, 83.8\% recall, and an F1-score of 84.1\%. These results underscore the complementary value of each modality and validate the effectiveness of multimodal fusion in enhancing enzyme function prediction.

\begin{table}[h]
    \caption{Comparison of Quantum Models for Enzyme Function Prediction}
    \renewcommand{\arraystretch}{1}
    \setlength{\tabcolsep}{2pt}
    \scriptsize
    \centering
    \begin{threeparttable}
    {\fontsize{9}{10}\selectfont
        \begin{tabular}{c|cccc}
            \hline
            Models & QSVM & QCNN & QGNN & \textbf{QVT} \\
            & & & & \small{\textbf{(Ours)}}\\
            \hline
            Accuracy (\%) & 74.5 & 78.3 & 81.7 & \textbf{85.1} \\
            MAC (Ops) & $1.2 \times 10^8$ & $2.0 \times 10^8$ & $3.8 \times 10^8$ & \textbf{$6.2 \times 10^8$} \\
            Circuit Depth & 12 & 16 & 20 & \textbf{24} \\
            Training Time (min) & 18 & 32 & 47 & \textbf{58} \\
            Memory (MB) & 45 & 110 & 180 & \textbf{240} \\
            \hline
        \end{tabular}}
        \begin{tablenotes}[flushleft]
            \item *All models were evaluated on the same enzyme classification dataset using identical hardware (Qiskit Aer simulator backend).
            \item *QVT incorporates multimodal biochemical encoders and cross-attention fusion.
        \end{tablenotes}
    \end{threeparttable}
    \label{tab:qml_comparison}
\end{table}

\autoref{tab:qml_comparison} presents a comparative analysis of four quantum machine learning models: QSVM, Quantum Convolutional Neural Network (QCNN), Quantum Graph Neural Network (QGNN), and the proposed QVT for the task of predicting enzyme function. The QVT model achieves the highest classification accuracy (85.1\%), outperforming all baselines, with incremental improvements observed across the QSVM (74.5\%), QCNN (78.3\%), and QGNN (81.7\%) models. In terms of computational complexity, QVT exhibits the highest Multiply-Accumulate (MAC) operations at $6.2 \times 10^8$, indicating a trade-off between predictive performance and resource requirements. This trend is also reflected in circuit depth, training time, and memory usage, where QVT incurs greater cost (24 layers, 58 minutes, and 240MB respectively) compared to the other models. All models were evaluated on a consistent dataset and quantum simulation backend (Qiskit), ensuring fair and controlled comparison.

\begin{figure}[ht]
\centering
\includegraphics[width=0.48\textwidth]{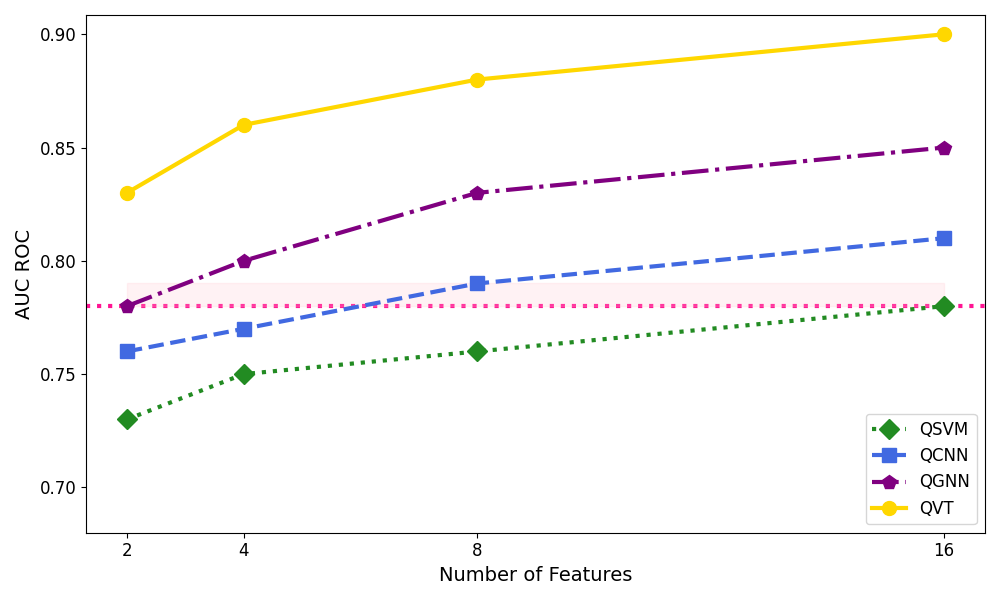}
\caption{Comparison of AUC-ROC performance across quantum models (QSVM, QCNN, QGNN, and QVT) under the Statevector backend. Results are shown for varying numbers of input enzyme features (2, 4, 8, 16). The QVT model consistently outperforms all other methods, with improvements becoming more pronounced as feature dimensionality increases. The horizontal pink band represents the baseline AUC (0.78) for tool reference.}
\label{fig:results}
\end{figure}

In the multimodal fusion layers, attention scores highlight correlations between electronic descriptors and graph substructures; for instance, regions of high nuclear repulsion energy tend to align with highly connected nodes in the molecular graph. These insights illustrate how QVT integrates data streams and enhances interpretability. To assess the distribution of misclassifications, a confusion matrix was generated for the top-1 predictions across six broad EC classes. Overall, the majority of errors involved closely related classes, such as oxidoreductases and lyases, which often share partial mechanistic overlap. In contrast, classes with more distinct catalytic mechanisms (isomerases versus ligases) exhibited fewer misclassifications. This pattern suggests that the QVT model captures class-specific biochemical signatures but can occasionally struggle with boundary cases where enzyme families share overlapping functional motifs. In practice, the strong performance on the top-1 accuracy measure indicates that the model ranks the correct enzyme function highly, even in these challenging scenarios.

\section{Conclusion} 
The QVT framework developed in this study demonstrates substantial improvements in enzyme function prediction through the integration of multiple biochemical modalities. Beginning with a baseline model that combines SMILES/SELFIES sequence embeddings and quantum chemical descriptors, achieving a Top-1 accuracy of 74.0\%, the successive addition of molecular graph structures, 2D molecular images, and molecular fingerprints progressively enhanced performance, culminating in an accuracy of 85.1\%. These results confirm that each modality captures distinct yet complementary biochemical properties, and their synergistic fusion leads to more accurate and robust enzyme classification.

\section{ Future Work} 
This study opens multiple promising avenues. First, we plan to extend our simulation-based evaluation to real quantum hardware by deploying the QVT model on IBM Quantum devices via the Qiskit runtime environment. Second, the graph representation pipeline will be extended to 3D conformer graphs using quantum geometry optimization, enabling finer spatial discrimination between isomeric enzymes. Third, focus on detailed investigations of specific enzyme families, such as DNA methyltransferases (DNMTs), by examining their distinct structural motifs, methylation patterns, and catalytic properties that influence targeted reactivity in single-enzyme contexts. This targeted approach would enable more refined parameterization of quantum descriptors for specialized proteins, highlighting how fine-drawn variations within active sites can result in significant functional differences.

\bibliographystyle{IEEEtran}
\bibliography{QCHEMNotes}

\begin{thebibliography}{10}
\providecommand{\url}[1]{#1}
\csname url@samestyle\endcsname
\providecommand{\newblock}{\relax}
\providecommand{\bibinfo}[2]{#2}
\providecommand{\BIBentrySTDinterwordspacing}{\spaceskip=0pt\relax}
\providecommand{\BIBentryALTinterwordstretchfactor}{4}
\providecommand{\BIBentryALTinterwordspacing}{\spaceskip=\fontdimen2\font plus
\BIBentryALTinterwordstretchfactor\fontdimen3\font minus \fontdimen4\font\relax}
\providecommand{\BIBforeignlanguage}[2]{{%
\expandafter\ifx\csname l@#1\endcsname\relax
\typeout{** WARNING: IEEEtran.bst: No hyphenation pattern has been}%
\typeout{** loaded for the language `#1'. Using the pattern for}%
\typeout{** the default language instead.}%
\else
\language=\csname l@#1\endcsname
\fi
#2}}
\providecommand{\BIBdecl}{\relax}
\BIBdecl

\bibitem{tripathi2020industrial}
P.~Tripathi and S.~Sinha, ``Industrial biocatalysis: An insight into trends and future directions,'' \emph{Current Sustainable/Renewable Energy Reports}, vol.~7, pp. 66--72, 2020.

\bibitem{shoda2016enzymes}
S.-i. Shoda, H.~Uyama, J.-i. Kadokawa, S.~Kimura, and S.~Kobayashi, ``Enzymes as green catalysts for precision macromolecular synthesis,'' \emph{Chemical reviews}, vol. 116, no.~4, pp. 2307--2413, 2016.

\bibitem{luo2025learning}
J.~Luo and Y.~Luo, ``Learning maximally spanning representations improves protein function annotation,'' in \emph{International Conference on Research in Computational Molecular Biology}.\hskip 1em plus 0.5em minus 0.4em\relax Springer, 2025, pp. 420--423.

\bibitem{thurimella2024harnessing}
K.~Thurimella, ``Harnessing deep learning with protein language models to unveil microbial enzyme function in health and disease,'' Ph.D. dissertation, 2024.

\bibitem{sampaio2023machine}
P.~S. Sampaio and P.~Fernandes, ``Machine learning: a suitable method for biocatalysis,'' \emph{Catalysts}, vol.~13, no.~6, p. 961, 2023.

\bibitem{salas2024machine}
L.~F. Salas-Nu{\~n}ez, A.~Barrera-Ocampo, P.~A. Caicedo, N.~Cortes, E.~H. Osorio, M.~F. Villegas-Torres, and A.~F. Gonz{\'a}lez~Barrios, ``Machine learning to predict enzyme--substrate interactions in elucidation of synthesis pathways: A review,'' \emph{Metabolites}, vol.~14, no.~3, p. 154, 2024.

\bibitem{catacutan2024machine}
D.~B. Catacutan, J.~Alexander, A.~Arnold, and J.~M. Stokes, ``Machine learning in preclinical drug discovery,'' \emph{Nature Chemical Biology}, vol.~20, no.~8, pp. 960--973, 2024.

\bibitem{kyro2025t}
G.~W. Kyro, A.~M. Smaldone, Y.~Shee, C.~Xu, and V.~S. Batista, ``T-alpha: A hierarchical transformer-based deep neural network for protein--ligand binding affinity prediction with uncertainty-aware self-learning for protein-specific alignment,'' \emph{Journal of Chemical Information and Modeling}, vol.~65, no.~5, pp. 2395--2415, 2025.

\bibitem{luo2022understanding}
S.~Luo, L.~Liu, C.-J. Lyu, B.~Sim, Y.~Liu, H.~Gong, Y.~Nie, and Y.-L. Zhao, ``Understanding the effectiveness of enzyme pre-reaction state by a quantum-based machine learning model,'' \emph{Cell Reports Physical Science}, vol.~3, no.~11, 2022.

\bibitem{ryu2019deep}
J.~Y. Ryu, H.~U. Kim, and S.~Y. Lee, ``Deep learning enables high-quality and high-throughput prediction of enzyme commission numbers,'' \emph{Proceedings of the National Academy of Sciences}, vol. 116, no.~28, pp. 13\,996--14\,001, 2019.

\bibitem{li2018deepre}
Y.~Li, S.~Wang, R.~Umarov, B.~Xie, M.~Fan, L.~Li, and X.~Gao, ``Deepre: sequence-based enzyme ec number prediction by deep learning,'' \emph{Bioinformatics}, vol.~34, no.~5, pp. 760--769, 2018.

\bibitem{biamonte2017quantum}
J.~Biamonte, P.~Wittek, N.~Pancotti, P.~Rebentrost, N.~Wiebe, and S.~Lloyd, ``Quantum machine learning,'' \emph{Nature}, vol. 549, no. 7671, pp. 195--202, 2017.

\bibitem{rupp2012fast}
M.~Rupp, A.~Tkatchenko, K.-R. M{\"u}ller, and O.~A. Von~Lilienfeld, ``Fast and accurate modeling of molecular atomization energies with machine learning,'' \emph{Physical review letters}, vol. 108, no.~5, p. 058301, 2012.

\bibitem{montavon2013machine}
G.~Montavon, M.~Rupp, V.~Gobre, A.~Vazquez-Mayagoitia, K.~Hansen, A.~Tkatchenko, K.-R. M{\"u}ller, and O.~A. Von~Lilienfeld, ``Machine learning of molecular electronic properties in chemical compound space,'' \emph{New Journal of Physics}, vol.~15, no.~9, p. 095003, 2013.

\bibitem{damborsky2024accelerating}
J.~Damborsky, P.~Kouba, J.~Sivic, D.~Bednar, and S.~Mazurenko, ``Accelerating enzyme engineering with quantum power: Quest for quantum advantage in biocatalysis,'' 2024.

\bibitem{mienye2024comprehensive}
I.~D. Mienye and T.~G. Swart, ``A comprehensive review of deep learning: Architectures, recent advances, and applications,'' \emph{Information}, vol.~15, no.~12, p. 755, 2024.

\bibitem{villalobos2022protein}
J.~Villalobos-Alva, L.~Ochoa-Toledo, M.~J. Villalobos-Alva, A.~Aliseda, F.~P{\'e}rez-Escamirosa, N.~F. Altamirano-Bustamante, F.~Ochoa-Fern{\'a}ndez, R.~Zamora-Sol{\'\i}s, S.~Villalobos-Alva, C.~Revilla-Monsalve \emph{et~al.}, ``Protein science meets artificial intelligence: a systematic review and a biochemical meta-analysis of an inter-field,'' \emph{Frontiers in Bioengineering and Biotechnology}, vol.~10, p. 788300, 2022.

\bibitem{gilmer2017neural}
J.~Gilmer, S.~S. Schoenholz, P.~F. Riley, O.~Vinyals, and G.~E. Dahl, ``Neural message passing for quantum chemistry,'' in \emph{International conference on machine learning}.\hskip 1em plus 0.5em minus 0.4em\relax PMLR, 2017, pp. 1263--1272.

\bibitem{coley2019graph}
C.~W. Coley, W.~Jin, L.~Rogers, T.~F. Jamison, T.~S. Jaakkola, W.~H. Green, R.~Barzilay, and K.~F. Jensen, ``A graph-convolutional neural network model for the prediction of chemical reactivity,'' \emph{Chemical science}, vol.~10, no.~2, pp. 370--377, 2019.

\bibitem{chandrasekaran2021image}
S.~N. Chandrasekaran, H.~Ceulemans, J.~D. Boyd, and A.~E. Carpenter, ``Image-based profiling for drug discovery: due for a machine-learning upgrade?'' \emph{Nature reviews drug discovery}, vol.~20, no.~2, pp. 145--159, 2021.

\bibitem{baltruvsaitis2018multimodal}
T.~Baltru{\v{s}}aitis, C.~Ahuja, and L.-P. Morency, ``Multimodal machine learning: A survey and taxonomy,'' \emph{IEEE transactions on pattern analysis and machine intelligence}, vol.~41, no.~2, pp. 423--443, 2018.

\bibitem{morgat2016updates}
A.~Morgat, T.~Lombardot, K.~B. Axelsen, L.~Aimo, A.~Niknejad, N.~Hyka-Nouspikel, E.~Coudert, M.~Pozzato, M.~Pagni, S.~Moretti \emph{et~al.}, ``Updates in rhea—an expert curated resource of biochemical reactions,'' \emph{Nucleic acids research}, p. gkw990, 2016.

\bibitem{pundir2017uniprot}
S.~Pundir, M.~J. Martin, and C.~O’Donovan, ``Uniprot protein knowledgebase,'' \emph{Protein Bioinformatics: From Protein Modifications and Networks to Proteomics}, pp. 41--55, 2017.

\bibitem{probst2022biocatalysed}
D.~Probst, M.~Manica, Y.~G. Nana~Teukam, A.~Castrogiovanni, F.~Paratore, and T.~Laino, ``Biocatalysed synthesis planning using data-driven learning,'' \emph{Nature communications}, vol.~13, no.~1, p. 964, 2022.

\bibitem{lo2023recent}
A.~Lo, R.~Pollice, A.~Nigam, A.~D. White, M.~Krenn, and A.~Aspuru-Guzik, ``Recent advances in the self-referencing embedded strings (selfies) library,'' \emph{Digital Discovery}, vol.~2, no.~4, pp. 897--908, 2023.

\bibitem{landrum2019rdkit}
G.~Landrum, ``{RDKit}: Open-source cheminformatics,'' \url{http://www.rdkit.org}, 2019.

\bibitem{Qchem}
{Q-Chem}, ``{Q-Chem}: Software for computing molecular electronic structures using quantum mechanics,'' \url{https://www.q-chem.com/}, 2019, accessed: 2025-08-09.

\bibitem{shao2015advances}
Y.~Shao, Z.~Gan, E.~Epifanovsky, A.~T. Gilbert, M.~Wormit, J.~Kussmann, A.~W. Lange, A.~Behn, J.~Deng, X.~Feng \emph{et~al.}, ``Advances in molecular quantum chemistry contained in the q-chem 4 program package,'' \emph{Molecular Physics}, vol. 113, no.~2, pp. 184--215, 2015.

\bibitem{gugler2022quantum}
S.~Gugler and M.~Reiher, ``Quantum chemical roots of machine-learning molecular similarity descriptors,'' \emph{Journal of Chemical Theory and Computation}, vol.~18, no.~11, pp. 6670--6689, 2022.

\bibitem{mottonen2004quantum}
M.~M{\"o}tt{\"o}nen, J.~J. Vartiainen, V.~Bergholm, and M.~M. Salomaa, ``Quantum circuits for general multiqubit gates,'' \emph{Physical review letters}, vol.~93, no.~13, p. 130502, 2004.

\bibitem{QiskitGitHub}
{Qiskit}, ``{Qiskit}: Open-source quantum computing framework,'' \url{https://github.com/Qiskit/qiskit}, 2019, accessed: 2025-08-09.

\end{thebibliography}

\end{document}